%% file: ijcai22.tex
\documentclass{article}
\usepackage{ijcai22}

\usepackage{times}
\usepackage{soul}
\usepackage{url}
\usepackage[hidelinks]{hyperref}
\usepackage[utf8]{inputenc}
\usepackage[small]{caption}
\usepackage{graphicx}
\usepackage{amsmath}
\usepackage{amsthm}
\usepackage{booktabs}
\usepackage{algorithm}
\usepackage{algorithmic}
\usepackage{multirow}
\usepackage{subfigure}
\usepackage{enumitem}
\usepackage{amssymb}

\title{Robust Weight Perturbation for Adversarial Training}

\author{
Chaojian Yu$^{1}$\footnote{This work is done during an internship at JD Explore Academy}
\and
Bo Han$^2$\and
Mingming Gong$^{3}$\and
Li Shen$^4$\and
Shiming Ge$^5$\and
Du Bo$^6$\and
Tongliang Liu$^{1}$\footnote{Corresponding author}
\affiliations
$^1$Trustworthy Machine Learning Lab, School of Computer Science, The University of Sydney, Australia\\
$^2$Department of Computer Science, Hong Kong Baptist University, China\\
$^3$School of Mathematics and Statistics, The University of Melbourne, Australia\\
$^4$JD Explore Academy, China\\
$^5$Institute of Information Engineering, Chinese Academy of Sciences, China\\
$^6$School of Computer Science, Wuhan University, China
\emails
\{chyu8051,tongliang.liu\}@sydney.edu.au,
bhanml@comp.hkbu.edu.hk,
mingming.gong@unimelb.edu.au,
mathshenli@gmail.com,
geshiming@iie.ac.cn,
gunspace@163.com
}

\begin{document}

\maketitle

\input{ijcai22_1_abstract}
\input{ijcai22_2_introduction}
\input{ijcai22_3_related_work}
\input{ijcai22_4_method}
\input{ijcai22_5_experiment}
\input{ijcai22_6_conclusion}

\section*{Acknowledgements}
This work is supported in part by Beijing Natural Science Foundation (19L2040), NSFC Young Scientists Fund No.~62006202, Guangdong Basic and Applied Basic Research Foundation No.~2022A1515011652, and Science and Technology Innovation 2030 –“Brain Science and Brain-like Research” Major Project (No. 2021ZD0201402 and No. 2021ZD0201405).

\clearpage
\bibliographystyle{named}
\bibliography{ijcai22}

\end{document}

%% file: ijcai22_1_abstract.tex
\begin{abstract}
  Overfitting widely exists in adversarial robust training of deep networks. An effective remedy is adversarial weight perturbation, which injects the worst-case weight perturbation during network training by maximizing the classification loss on adversarial examples. Adversarial weight perturbation helps reduce the robust generalization gap; however, it also undermines the robustness improvement. A criterion that regulates the weight perturbation is therefore crucial for adversarial training. In this paper, we propose such a criterion, namely Loss Stationary Condition (LSC) for constrained perturbation. With LSC, we find that it is essential to conduct weight perturbation on adversarial data with small classification loss to eliminate robust overfitting. Weight perturbation on adversarial data with large classification loss is not necessary and may even lead to poor robustness. Based on these observations, we propose a robust perturbation strategy to constrain the extent of weight perturbation. The perturbation strategy prevents deep networks from overfitting while avoiding the side effect of excessive weight perturbation,  significantly improving the robustness of adversarial training. Extensive experiments demonstrate the superiority of the proposed method over the state-of-the-art adversarial training methods.
\end{abstract}

%% file: ijcai22_2_introduction.tex
\section{Introduction}
Although deep neural networks (DNNs) have led to impressive breakthroughs in a number of fields such as computer vision~\cite{he2016deep}, speech recognition~\cite{wang2017residual}, and NLP~\cite{devlin2018bert}, they are extremely vulnerable to adversarial examples that are crafted by adding small and human-imperceptible perturbation to normal examples~\cite{szegedy2013intriguing,goodfellow2014explaining}. 

The vulnerability of DNNs has attracted extensive attention and led to a large number of defense techniques against adversarial examples. Across existing defenses, adversarial training (AT) is one of the strongest empirical defenses. AT directly incorporates adversarial examples into the training process to solve a min-max optimization problem~\cite{madry2017towards}, which can obtain models with moderate adversarial robustness and has not been comprehensively attacked~\cite{athalye2018obfuscated}. However, different from the natural training scenario, overfitting is a dominant phenomenon in adversarial robust training of deep networks~\cite{rice2020overfitting}. After a certain point in AT, the robust performance on test data will continue to degrade with further training, as shown in Figure \ref{fig:1}(a). This phenomenon, termed as \emph{robust overfitting}, breaches the common practice in deep learning that using over-parameterized networks and training for as long as possible~\cite{belkin2019reconciling}. Such anomaly in AT causes detrimental effects on the robust generalization performance and subsequent algorithm assessment~\cite{rice2020overfitting,chen2020robust}. Relief techniques that mitigate robust overfitting have thus become crucial for adversarial training.

An effective remedy for robust overfitting is Adversarial Weight Perturbation (AWP)~\cite{wu2020adversarial}, which forms a double-perturbation mechanism
that adversarially perturbs both inputs and weights:
\begin{equation}\label{eq1}
    \min_w \max_{v \in \mathcal{V}} \frac{1}{n} \sum_{i=1}^{n} \max_{||x_{i}' - x_i||_p \le \epsilon} \ell(f_{w+v}(x_{i}'),y_{i}),
\end{equation}
where $n$ is the number of training examples, $x_i'$ is the adversarial example of $x_i$, $f_w$ is the DNN with weight $w$, $\ell(\cdot)$ is the loss function, $\epsilon$ is the maximum perturbation constraint for inputs (\textit{i.e.}, $||x_{i}' - x_i||_p \le \epsilon$), and $\mathcal{V}$ is the feasible perturbation region for weights (\textit{i.e.}, $\{v \in \mathcal{V} : ||v||_2 \le \gamma||w||_2\}$, where $\gamma$ is the constraint on weight perturbation size). The inner maximization is to find adversarial examples $x_i'$ within the $\epsilon$-ball centered at normal examples $x_i$ that maximizes the classification loss $\ell$. On the other hand, the outer maximization is to find weight perturbation $v$ that maximizes the loss $\ell$ on adversarial examples to 
reduce robust generalization gap. This is the problem of training a weight-perturbed robust classifier on adversarial examples. Therefore, how well the weight perturbation is found directly affects the performance of the outer minimization, \textit{i.e.}, the robustness of the classifier.

Several attack methods have been used to solve the inner maximization problem in Eq.(\ref{eq1}), such as Fast Gradient Sign Method (FGSM)~\cite{goodfellow2014explaining} and Projected Gradient Descent (PGD)~\cite{madry2017towards}. 
For the outer maximization problem, AWP~\cite{wu2020adversarial} injects the worst-case weight perturbation to reduce robust generalization gap. However, the extent to which the weights should be perturbed has not been explored. Without an appropriate criterion to regulate the weight perturbation, the adversarial training procedure is difficult to unleash its full power, since worst-case weight perturbation will undermine the robustness improvement (in Section \ref{section:3}). In this paper, we propose such a criterion, namely Loss Stationary Condition (LSC) for constrained perturbation (in Section \ref{section:3}), which helps to better understand robust overfitting, and this in turn motivates us to propose an improved weight perturbation strategy for better adversarial robustness (in Section \ref{section:4}).
Our main contributions as follows:
\begin{itemize}[leftmargin=*]
\item We propose a principled criterion LSC to analyse the adversarial weight perturbation. It provides a better understanding of robust overfitting in adversarial training, and it is also a good indicator for efficient weight perturbation.
\item With LSC, we find that better perturbation of model weights is associated with perturbing on adversarial data with small classification loss. For adversarial data with large classification loss, weight perturbation is not necessary and can even be harmful.
\item We propose a robust perturbation strategy to constrain the extent of weight perturbation. Experiments show that the robust strategy significantly improves the robustness of adversarial training.
\end{itemize}

%% file: ijcai22_3_related_work.tex
\section{Related Work}
\subsection{Adversarial Attacks}
Let $\mathcal{X}$ denote the input feature space and $\mathcal{B}_{\epsilon}^{p}(x) = \{x' \in \mathcal{X}:||x'-x||_p \le \epsilon\}$ be the $\ell_p$-norm ball of radius $\epsilon$ centered at $x$ in $\mathcal{X}$. Here we selectively introduce several commonly used adversarial attack methods.

\paragraph{Fast Gradient Sign Method (FGSM).} FGSM \cite{goodfellow2014explaining} perturbs natural example $x$ for one step with step size $\epsilon$ along the gradient direction:
\begin{equation}
    x'=x+\epsilon \cdot \mathrm{sign}(\nabla_x\ell(f_w(x),y)).
\end{equation}

\paragraph{Projected Gradient Descent (PGD).} PGD \cite{madry2017towards} is a stronger iterative variant of FGSM, which perturbs normal example $x$ for multiple steps $K$ with a smaller step size $\alpha$:
\begin{equation}
    x^0 \sim \mathcal{U}(\mathcal{B}_{\epsilon}^{p}(x)),
\end{equation}
\begin{equation}
    x^k=\Pi_{\mathcal{B}_{\epsilon}^{p}(x)}(x^{k-1}+\alpha \cdot \mathrm{sign}(\nabla_{x^{k-1}}\ell(f_w(x^{k-1}),y))),
\end{equation}
where $\mathcal{U}$ denotes the uniform distribution, $x^0$ denotes the normal example disturbed by a small uniform random noise, $x^k$ denotes the adversarial example at step $k$, and $\Pi_{\mathcal{B}_{\epsilon}^{p}(x)}$ denotes the projection function that projects the adversarial example back into the set $\mathcal{B}_{\epsilon}^{p}(x)$ if necessary.

\paragraph{AutoAttack (AA).} AA \cite{croce2020reliable} is an ensemble of complementary attacks, which consists of three white-box attacks and a black-box attack. AA regards models to be robust only if the models correctly classify all types of adversarial examples, which is among the most reliable evaluation of adversarial robustness to date.

\subsection{Adversarial Defense}
Since the discovery of adversarial examples, a large number of works have emerged for defending against adversarial attacks, such as input denoising \cite{wu2021attacking}, modeling adversarial noise \cite{zhou2021modeling}, and adversarial training \cite{goodfellow2014explaining,madry2017towards}. Among them, adversarial training has been demonstrated to be one of the most effective method \cite{athalye2018obfuscated}. Based on adversarial training, a wide range of subsequent works are then proposed to further improve the model robustness. Here, we introduce two currently state-of-the-art AT frameworks.

\paragraph{TRADES.} TRADES \cite{zhang2019theoretically} optimizes a regularized surrogate loss that is a trade-off between the natural accuracy and adversarial robustness:
\begin{align}
    \ell^{\mathrm{TRADES}}&(w;x,y)  = \frac{1}{n} \sum_{i=1}^{n} \big\{\mathrm{CE}(f_{w}(x_i),y_i) \nonumber\\
    & + \beta \cdot \max_{x' \in \mathcal{B}_{\epsilon}^{p}(x)} \mathrm{KL}(f_{w}(x_i)||f_{w}(x_i')) \big\},
\end{align}
where CE is the cross-entropy loss that encourages the network to maximize the natural accuracy, KL is the Kullback-Leibler divergence that encourages to improve the robust accuracy, and $\beta$ is the hyperparameter to control the trade-off between natural accuracy and adversarial robustness.

\paragraph{Robust Self-Training (RST).} RST \cite{carmon2019unlabeled} utilize additional 500K unlabeled data extracted from the 80 Million Tiny Images dataset. RST first leverages the surrogate natural model to generate pseudo-labels for these unlabeled data, and then adversarially trains the network with both additional pseudo-labeled unlabeled data $(\tilde{x},\tilde{y})$ and original labeled data $(x,y)$ in a supervised setting:
\begin{equation}
\resizebox{.91\linewidth}{!}{$
    \ell^{\mathrm{RST}}(w;x,y,\tilde{x},\tilde{y})=\ell^{\mathrm{TRADES}}(w;x,y) + \lambda \cdot \ell^{\mathrm{TRADES}}(w;\tilde{x},\tilde{y}),
$}
\end{equation}
where $\lambda$ is the weight on unlabeled data.

\subsection{Robust Overfitting}

\begin{figure*}[!t]
\centering
    \subfigure[Learning curve]{
        \includegraphics[width=0.46\columnwidth]{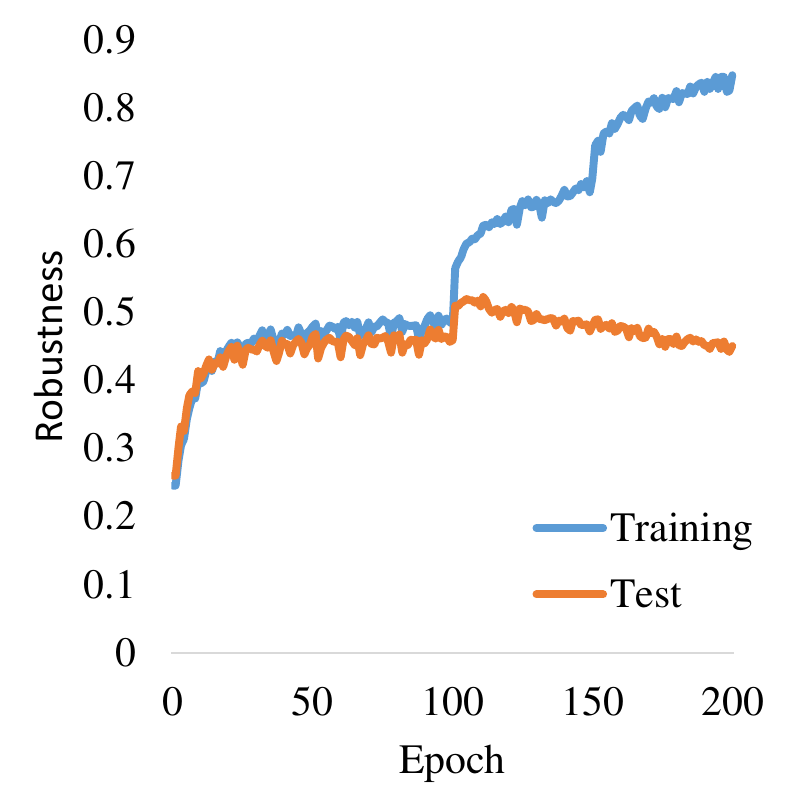}
    }
    \subfigure[Robustness vs. Weight perturbation size]{
        \includegraphics[width=0.46\columnwidth]{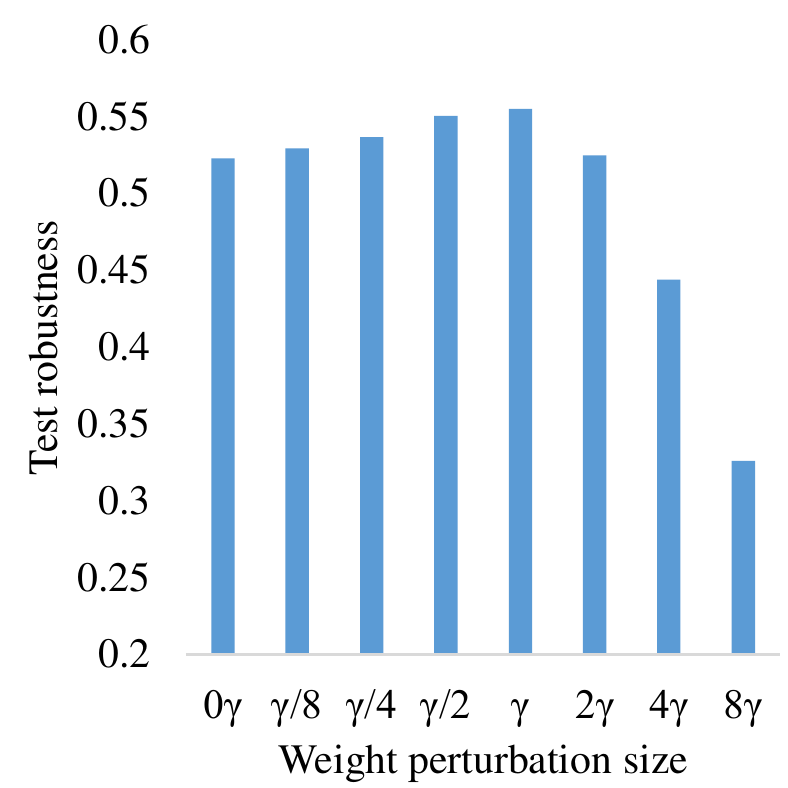}
        \includegraphics[width=0.46\columnwidth]{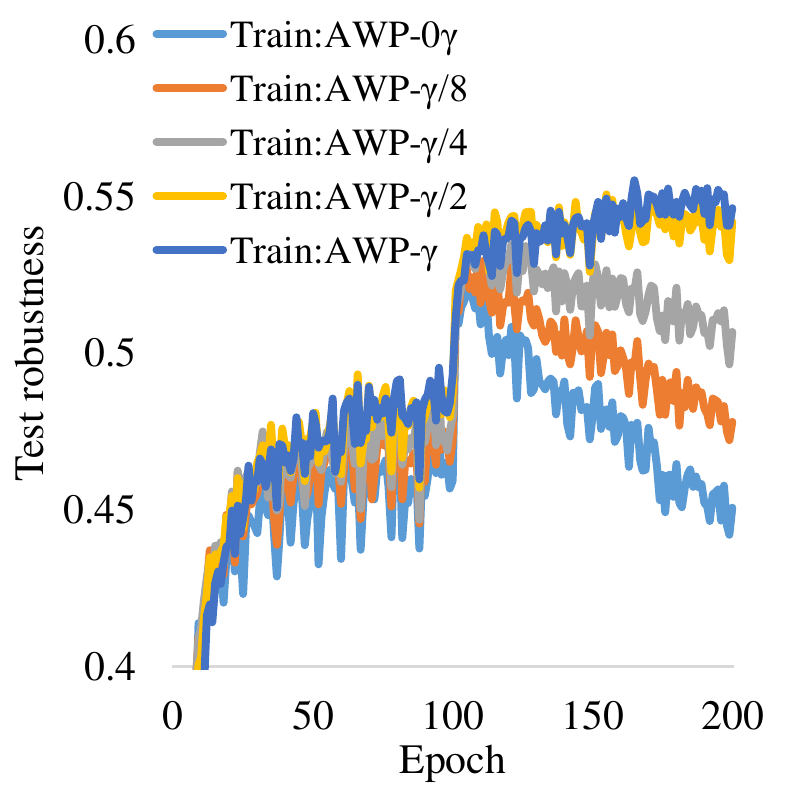}
        \includegraphics[width=0.46\columnwidth]{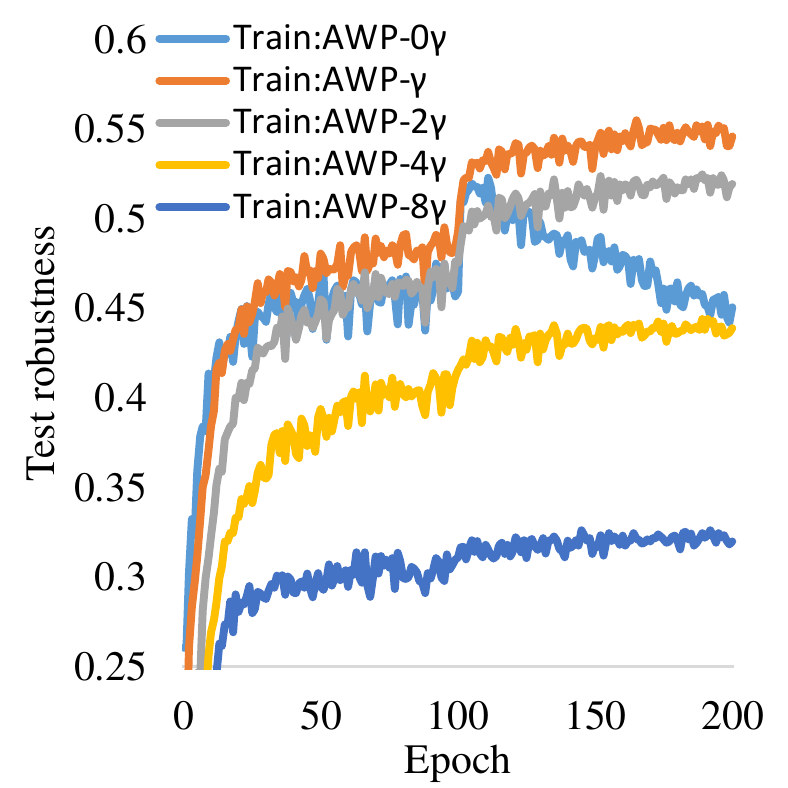}
    }
    \\
    \subfigure[Robustness vs. LSC range]{
        \includegraphics[width=0.46\columnwidth]{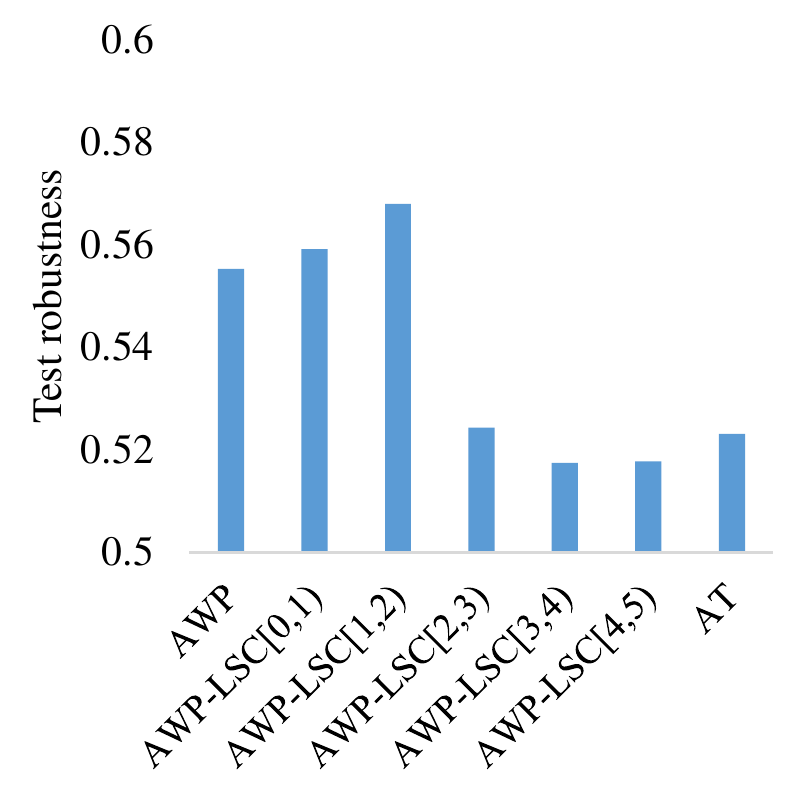}
        \includegraphics[width=0.46\columnwidth]{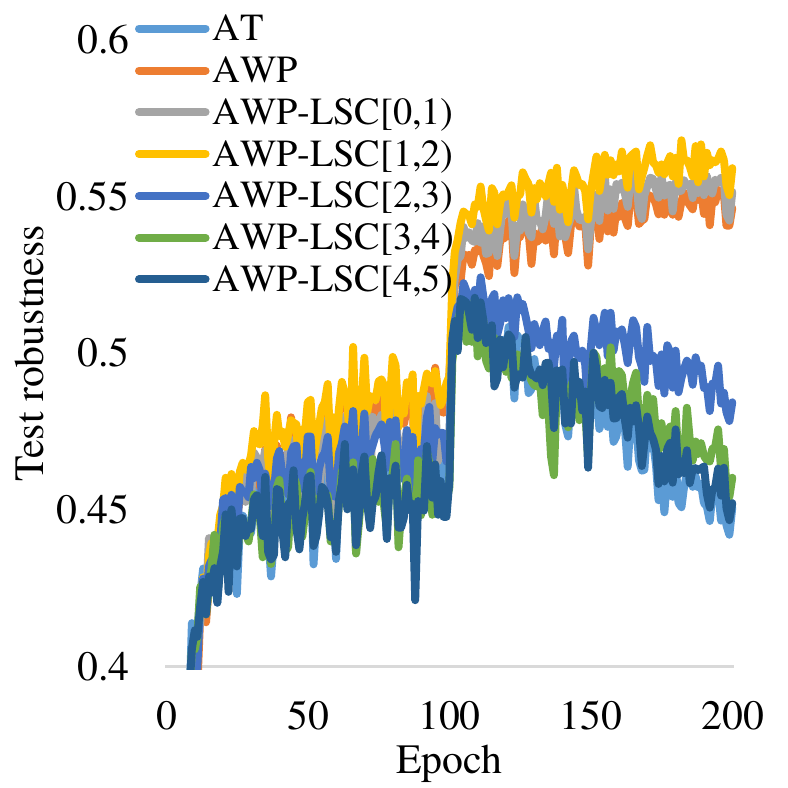}
        \includegraphics[width=0.46\columnwidth]{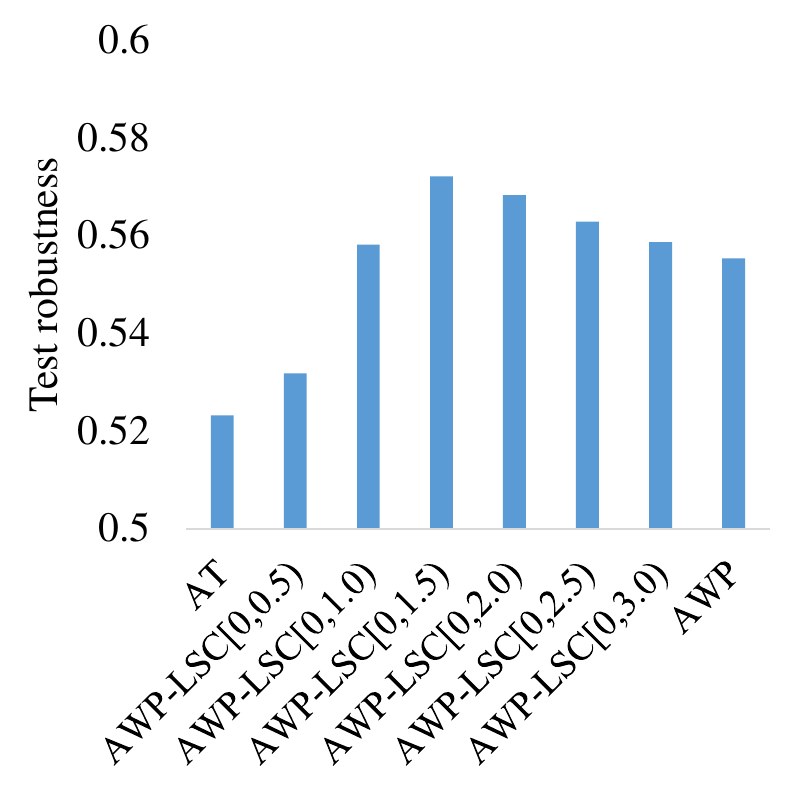}
        \includegraphics[width=0.46\columnwidth]{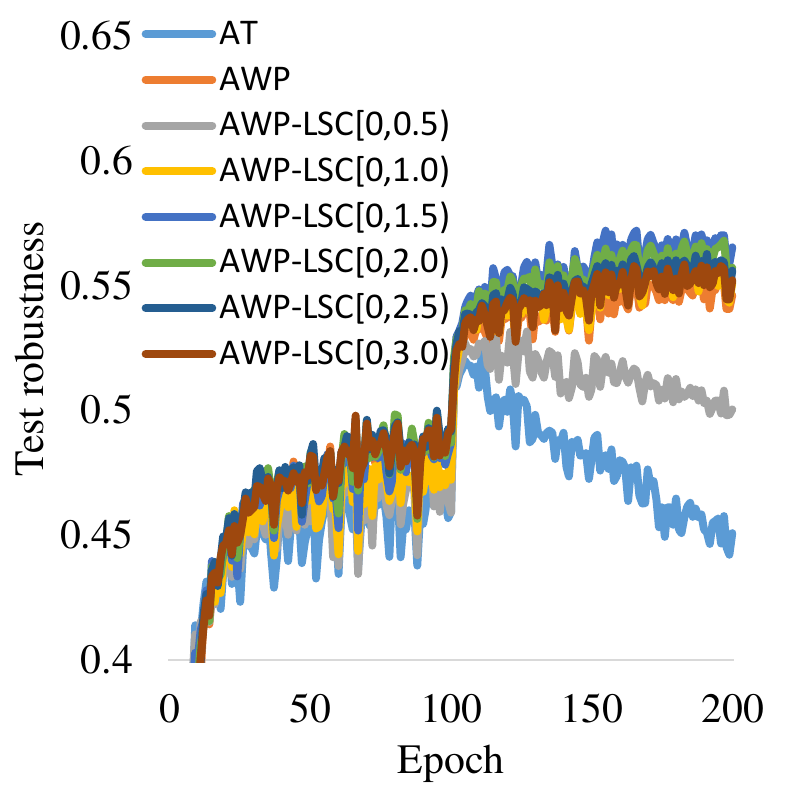}
    }
\caption{(a): The learning curve of vanilla AT; (b): Test robustness of AWP with varying weight perturbation size; (c): Test robustness of AWP with varying LSC range.}
\label{fig:1}
\end{figure*}

Nowadays, there are effective countermeasures to alleviate the overfitting in natural training. But in adversarial training, robust overfitting widely exists and those common countermeasures used in natural training help little~\cite{rice2020overfitting}. \cite{schmidt2018adversarially} explains robust overfitting partially from the perspective of sample complexity, and is supported by empirical results in derivative works, such as adversarial training with semi-supervised learning \cite{carmon2019unlabeled,uesato2019labels,zhai2019adversarially}, robust local feature \cite{song2019robust} and data interpolation \cite{zhang2019adversarial,lee2020adversarial,chen2021guided}. Separate works have also attempt to mitigate robust overfitting by the unequal treatment of data \cite{zhang2020geometry} and weight smoothing \cite{chen2020robust}. Recent study \cite{wu2020adversarial} reveals the connection between the flatness of weight loss landscape and robust generalization gap, and proposes to incorporate adversarial weight perturbation mechanism in the adversarial training framework.
Despite the efficacy of adversarial weight perturbation in suppressing the robust overfitting, a deeper understanding of robust overfitting and a clear direction for valid weight perturbation is largely missing. The outer maximization in Eq.(\ref{eq1}) lacks an effective criterion to regulate and constrain the extent of weight perturbation, which in turn influences the optimization of the outer minimization. In this paper, we propose such a criterion and provide new understanding of robust overfitting. Following this,  we design a robust weight perturbation strategy that significantly improves the robustness of adversarial training.

%% file: ijcai22_4_method.tex
\section{Loss Stationary Condition}
\label{section:3}

In this section, we first empirically investigate the relationship between weight perturbation robustness and adversarial robustness, and then propose a criterion to analyse the adversarial weight perturbation, which leads to a new perspective of robust overfitting. To this end, some discussions about robust overfitting and weight perturbation are provided.

\paragraph{Does Weight Perturbation Robustness Certainly Lead to Better Adversarial Robustness?} First, we investigate whether the robustness against weight perturbation is beneficial to the adversarial robustness. In particular, we train PreAct ResNet-18 with AWP on CIFAR-10 using varying weight perturbation size from $0\gamma$, $\gamma/8$, $\gamma/4$, $\gamma/2$, $\gamma$, $2\gamma$, $4\gamma$ to $8\gamma$. In each setting, we evaluate the robustness of the model against 20-step PGD (PGD-20) attacks on CIFAR-10 test images. As shown in Figure \ref{fig:1}(b), when weight perturbation size is small, the best adversarial robustness has a certain improvement. However, when weight perturbation size is large, the best adversarial robustness begins to decrease significantly as the size of the perturbation increases. It can be explained by the fact that the network has to sacrifice adversarial robustness to allocate more capacity to defend against weight perturbation when weight perturbation size is large, which indicates that weight perturbation robustness and adversarial robustness are not mutually beneficial. As shown in Figure \ref{fig:1}(b), the performance gain of AWP is mainly due to suppressing robust overfitting.
 
\paragraph{Loss Stationary Condition.} In order to further analyse the weight perturbation, we propose a criterion that divides the training adversarial examples into different groups according to their classification loss:
\begin{equation}
\mathrm{LSC}[p,q] = \{x' \in \mathcal{X}~|~p \le \ell(f_w(x'),y) \le q\},
\end{equation}
where $p \le q$. The adversarial data in the group all satisfy their adversarial loss within a certain range, which is termed Loss Stationary Condition (LSC). The proposed criterion LSC allows the analysis of grouped adversarial data independently, and provides more insights into the robust overfitting.

\paragraph{LSC view of Adversarial Weight Perturbation.} To provide more insight into how AWP suppresses robust overfitting, we train PreAct ResNet-18  on CIFAR-10 by varying the LSC group that performs adversarial weight perturbation. In each setting, we evaluate the robustness of the model against PGD-20 attacks on CIFAR-10 test images. As shown in Figure \ref{fig:1}(c), when varying the LSC range, we can observe that conducting adversarial weight perturbation on adversarial examples with small classification loss is sufficient to eliminate robust overfitting. However, conducting adversarial weight perturbation on adversarial examples with large classification loss fails to suppress robust overfitting. The results indicate that to eliminate robust overfitting, it is essential to prevent the model from memorizing these easy-to-learn adversarial examples.
Besides, it is observed that conducting adversarial weight perturbation on adversarial examples with large classification loss leads to worse adversarial robustness, which again verifies that the robustness against weight perturbation will not bring adversarial robustness gain, or even on the contrary, it undermines the adversarial robustness enhancement.

\paragraph{Do We Really Need the Worst-case Weight Perturbation?} As aforementioned, the robustness against weight perturbation is not beneficial to the adversarial robustness improvement. Therefore, to purely eliminate robust overfitting, conducting worst-case weight perturbation on these adversarial examples is not necessary. In the next section, we will propose a robust perturbation strategy to address this issue.

\begin{algorithm}[t]
   \caption{Robust Weight Perturbation (RWP)}
   \label{alg:1}
\begin{algorithmic}
   \STATE {\bfseries Input:} Network $f_w$, training data $S$, mini-batch $\mathcal{B}$, batch size $n$, learning rate $\eta$, PGD step size $\alpha$, PGD steps $K_1$, PGD constraint $\epsilon$, RWP steps $K_2$, RWP constraint $\gamma$, minimum loss value $c_{min}$.
   \STATE {\bfseries Output:} Adversarially robust model $f_w$.
   \REPEAT
   \STATE Read mini-batch $x_{\mathcal{B}}$ from training set $S$.
   \STATE $x_{\mathcal{B}}' \leftarrow x_{\mathcal{B}} + \delta$, where $\delta \sim \mathrm{Uniform}(-\epsilon,\epsilon)$
   \FOR{$k=1$ {\bfseries to} $K_1$}
   \STATE $x_{\mathcal{B}}' \leftarrow \Pi_{\epsilon}(x_{\mathcal{B}}'+\alpha \cdot \mathrm{sign}(\nabla_{x_{\mathcal{B}}'}\ell(f_w(x_{\mathcal{B}}'),y)))$
   \ENDFOR
   \STATE Initialize $v = \mathbf{0}$
   \FOR{$k=1$ {\bfseries to} $K_2$}
   \STATE $V = \mathbb{I}_{\mathcal{B}}(\ell(f_{w+v}(x_{\mathcal{B}}'),y) \le c_{min})$
   \IF{$\sum{V} = 0$}
   \STATE {\bfseries break}
   \ELSE
   \STATE $v \leftarrow v + \nabla_{v}(V \cdot \ell(f_{w+v}(x_{\mathcal{B}}'),y))$
   \STATE $v \leftarrow \gamma\frac{v}{||v||}||w||$
   \ENDIF
   \ENDFOR
   \STATE $w \leftarrow (w + v) - \eta\nabla_{w+v}\frac{1}{n}\sum_{i=1}^{n}\ell(f_{w+v}({x'}_{\mathcal{B}}^{(i)}),y^{(i)}) -v$
   \UNTIL{training converged}
\end{algorithmic}
\end{algorithm}

\section{Robust Weight Perturbation}
\label{section:4}
As mentioned in Section \ref{section:3}, conducting adversarial weight perturbation on adversarial examples with small classification loss is enough to prevent robust overfitting and leads to higher robustness. However, conducting adversarial weight perturbation on adversarial examples with large classification loss may not be helpful. Recalling the criterion LSC proposed in Section \ref{section:3}, we have seen that the loss is closely correlated with the tendency of adversarial example to be overfitted. Thus, it can be used to constrain the extent of weight perturbation at a fine-grained level. Therefore, we propose to conduct weight perturbation on adversarial examples that are below a minimum loss value, so as to ensure that no robust overfitting occurs while avoiding the side effect of excessive weight perturbation.
Let $c_{min}$ be the minimum loss value. Instead of generating weight perturbation $v$ via outer maximization in Eq.(\ref{eq1}), we generate $v$ as follows:
\begin{equation}
\begin{split}
v^{k+1} = v^{k} + \nabla_{v^{k}}\frac{1}{n}\sum_{i=1}^{n} \mathbb{I}(x_{i}',y_i) \ell(f_{w+v^{k}}(x_{i}'),y_i), \\
\mathrm{where}~~~~ 
\mathbb{I}(x_{i}',y_i) =
\left\{
\begin{aligned}
0 & ~~~~\mathrm{if}~~ \ell(f_{w+v^{k}}(x_{i}'),y_i) > c_{min} \\
1 & ~~~~\mathrm{if}~~ \ell(f_{w+v^{k}}(x_{i}'),y_i) \le c_{min}\\
\end{aligned}
\right.
\end{split}
\end{equation}

The proposed Robust Weight Perturbation (RWP) algorithm is shown in Algorithm 1. We use PGD attack \cite{madry2017towards} to generate the training adversarial examples, which can be also extended to other variants such as TRADES \cite{zhang2019theoretically} and RST \cite{carmon2019unlabeled}. The mimimum loss value $c_{min}$ controls the extent of weight perturbation during network training. For example, in the early stages of training, the classification loss of adversarial example is generally larger than $c_{min}$ corresponding to no weight perturbation process. The classification loss of adversarial examples then decreases as training progresses. At each optimization step, we monitor the classification loss of the adversarial example and conduct the weight perturbation process for adversarial examples whose classification loss is smaller than $c_{min}$, enabled by an indicator control vector $V$. At each perturbation step, the weight perturbation $v$ will be updated to increase the classification loss of the corresponding adversarial example. When the classification loss of training adversarial examples is all higher than $c_{min}$ or the number of perturbation step reaches the defined value, we stop the weight perturbation process and inject the generated weight perturbation $v$ for adversarial training.

%% file: ijcai22_5_experiment.tex
\section{Experiments}

\begin{table*}[h]
\small
  \centering
  \begin{tabular}{clcccccccc}
    \toprule
    \multirow{2}*{Threat Model} & \multirow{2}*{Method} & \multicolumn{2}{c}{SVHN} & & \multicolumn{2}{c}{CIFAR-10} & & \multicolumn{2}{c}{CIFAR-100} \\
    \cmidrule{3-4}
    \cmidrule{6-7}
    \cmidrule{9-10}
    & & Best & Last & & Best & Last & & Best & Last \\
    \midrule
    \multirow{3}*{$L_\infty$} & AT & 53.22 $\pm$ 0.20 & 45.13 $\pm$ 0.17 & & 52.32 $\pm$ 0.31 & 45.08$\pm$ 0.19 & & 27.79$\pm$ 0.45 & 20.95$\pm$ 0.30\\
     & AT-AWP & 59.49$\pm$ 0.15 & 55.16$\pm$ 0.10 & & 55.54$\pm$ 0.20 & 54.64$\pm$ 0.25 & & 30.89$\pm$ 0.21 & 30.48$\pm$ 0.43\\
     & AT-RWP & \textbf{61.15$\pm$ 0.16} & \textbf{57.45$\pm$ 0.23} & & \textbf{58.55$\pm$ 0.50} & \textbf{58.01$\pm$ 0.33} & & \textbf{31.17$\pm$ 0.18} & \textbf{30.64$\pm$ 0.24} \\
    \midrule
    \multirow{3}*{$L_2$} & AT & 66.71$\pm$ 0.24 & 65.25$\pm$ 0.19 & & 69.40$\pm$ 0.38 & 66.02$\pm$ 0.15 & & 40.95$\pm$ 0.13 & 36.24$\pm$ 0.26\\
     & AT-AWP & 72.80$\pm$ 0.30 & 68.40$\pm$ 0.20 & & 72.72$\pm$ 0.21 & 72.48$\pm$ 0.45 & & 45.63$\pm$ 0.48 & 44.98$\pm$ 0.30\\
     & AT-RWP & \textbf{73.35$\pm$ 0.20} & \textbf{69.48$\pm$ 0.32} & & \textbf{74.47$\pm$ 0.14} & \textbf{73.84$\pm$ 0.27} & & \textbf{45.71$\pm$ 0.17} & \textbf{45.05$\pm$ 0.30} \\
    \bottomrule
  \end{tabular}
  \caption{Test robustness (\%) of AT, AT-AWP and AT-RWP using PreAct ResNet-18.}
  \label{table:1}
\end{table*}

\begin{table*}[!t]
\small
  \centering
  \begin{tabular}{lc|cccc|c}
    \toprule
    Defense & Natural & FGSM & PGD-20 & PGD-100 & C\&W$_\infty$ & AA\\
    \midrule
    AT & 86.52$\pm$ 0.57 & 61.91$\pm$ 0.15 & 55.47$\pm$ 0.10 & 55.15$\pm$ 0.28 & 54.51$\pm$ 0.19 & 52.18$\pm$ 0.04\\
    AT-AWP & 85.67$\pm$ 0.40 & 64.31$\pm$ 0.23 & 58.57$\pm$ 0.22 & 58.46$\pm$ 0.17 & 55.78$\pm$ 0.32 & 53.63$\pm$ 0.09\\
    AT-RWP & \textbf{86.86$\pm$ 0.51} & \textbf{66.22$\pm$ 0.31} & \textbf{62.87$\pm$ 0.25} & \textbf{62.87$\pm$ 0.34} & \textbf{56.62$\pm$ 0.18} & \textbf{54.61$\pm$ 0.11}\\
    \midrule
    TRADES & 84.42$\pm$ 0.36 & 61.20$\pm$ 0.09 & 56.05$\pm$ 0.13 & 55.85$\pm$ 0.20 & 53.67$\pm$ 0.14 & 52.64$\pm$ 0.07\\
    TRADES-AWP & 84.55$\pm$ 0.30 & 62.99$\pm$ 0.30 & 59.20$\pm$ 0.24 & 59.05$\pm$ 0.31 & 55.92$\pm$ 0.20 & 55.32$\pm$ 0.05\\
    TRADES-RWP & \textbf{86.14$\pm$ 0.43} & \textbf{64.70$\pm$ 0.17} & \textbf{60.45$\pm$ 0.19} & \textbf{60.30$\pm$ 0.30} & \textbf{58.07$\pm$ 0.33} & \textbf{57.20$\pm$ 0.09}\\
    \midrule
    RST & \textbf{89.88$\pm$ 0.36} & \textbf{70.08$\pm$ 0.62} & 62.40$\pm$ 0.51 & 62.08$\pm$ 0.31 & 61.14$\pm$ 0.46 & 59.71$\pm$ 0.10\\
    RST-AWP & 88.01$\pm$ 0.68 & 68.00$\pm$ 0.23 & 63.67$\pm$ 0.38 & 63.50$\pm$ 0.11 & 60.55$\pm$ 0.21 & 59.80$\pm$ 0.08\\
    RST-RWP & 88.87$\pm$ 0.55 & 69.71$\pm$ 0.12 & \textbf{64.11$\pm$ 0.16} & \textbf{63.92$\pm$ 0.26} & \textbf{62.03$\pm$ 0.23} & \textbf{60.36$\pm$ 0.06} \\
    \bottomrule
  \end{tabular}
  \caption{Test robustness (\%) on CIFAR-10 using Wide ResNet under $L_\infty$ threat model.}
  \label{table:2}
\end{table*}

In this section, we conduct comprehensive experiments to evaluate the effectiveness of RWP including its experimental settings, robustness evaluation and ablation studies.
\subsection{Experimental Setup}
\label{experiment setup}

\paragraph{Baselines and Implementation Details.} Our implementation is based on PyTorch and the code is publicly available\footnote{\url{https://github.com/ChaojianYu/Robust-Weight-Perturbation}}. We conduct extensive experiments across three benchmark datasets (CIFAR-10, CIFAR-100 and SVHN) and two threat models ($L_\infty$ and $L_2$). We use PreAct ResNet-18 \cite{he2016deep} and Wide ResNet (WRN-28-10 and WRN-34-10) \cite{zagoruyko2016wide} as the network structure following \cite{wu2020adversarial}. We compare the performance of the proposed method on a number of baseline methods: 1) standard adversarial training without weight perturbation, including vanilla AT \cite{madry2017towards}, TRADES \cite{zhang2019theoretically} and RST \cite{carmon2019unlabeled}; 2) adversarial training with AWP \cite{wu2020adversarial}, including AT-AWP, TRADES-AWP and RST-AWP. For training, the network is trained for 200 epochs using SGD with momentum 0.9, weight decay $5\times 10^{-4}$, and an initial learning rate of 0.1. The learning rate is divided by 10 at the 100-th and 150-th epoch. Standard data augmentation including random crops with 4 pixels of padding and random horizontal flips are applied. For testing, model robustness is evaluated by measuring the accuracy of the model under different adversarial attacks. For hyper-parameters in RWP, we set perturbation step $K_2=10$ for all datasets. The minimum loss value $c_{min}=1.7$ for CIFAR-10 and SVHN, and $c_{min}=4.0$ for CIFAR-100. The weight perturbation budget of $\gamma=0.01$ for AT-RWP, $\gamma=0.005$ for TRADES-RWP and RST-RWP following literature \cite{wu2020adversarial}. Other hyper-parameters of the baselines are configured as per their original papers. 

\paragraph{Adversarial Setting.} The training attack is 10-step PGD attack with random start. We follow the same settings in \cite{rice2020overfitting} : for $L_\infty$ threat model, $\epsilon=8/255$, step size $\alpha=1/255$ for SVHN, and $\alpha=2/255$ for both CIFAR10 and CIFAR100; for $L_2$ threat model, $\epsilon=128/255$, step size $\alpha=15/255$ for all datasets, which is a standard setting for adversarial training \cite{madry2017towards}. 
The test attacks used for robustness evaluation contains FGSM, PGD-20, PGD-100, C\&W$_\infty$ and Auto Attack (AA).

\begin{figure*}[t]
\centering
    \subfigure[Importance of minimum loss value $c_{min}$]{
        \includegraphics[width=0.6\columnwidth]{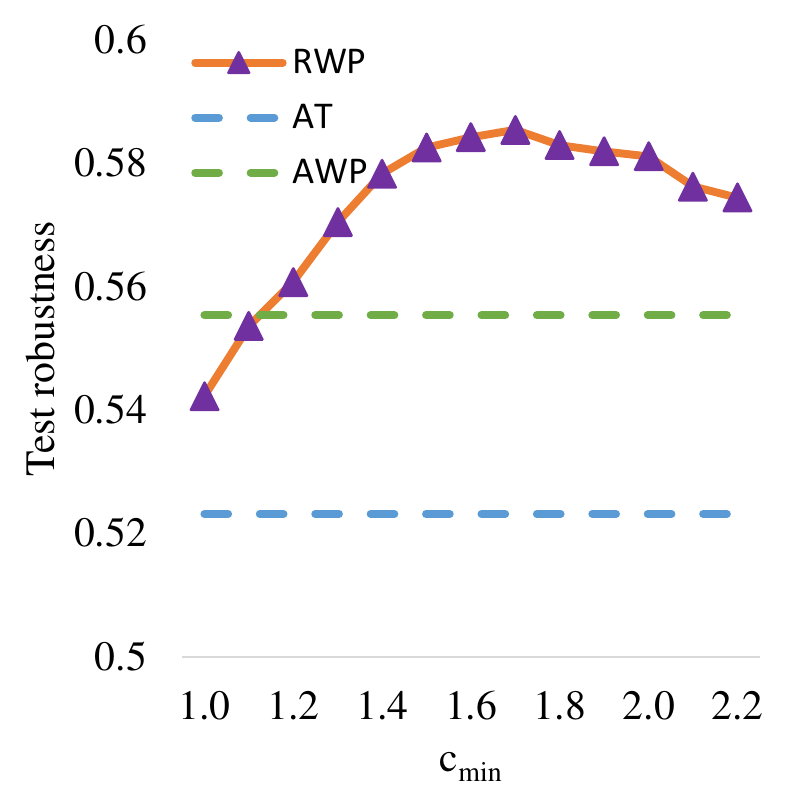}
    }
    \subfigure[Impacts of step number $K_2$]{
        \includegraphics[width=0.6\columnwidth]{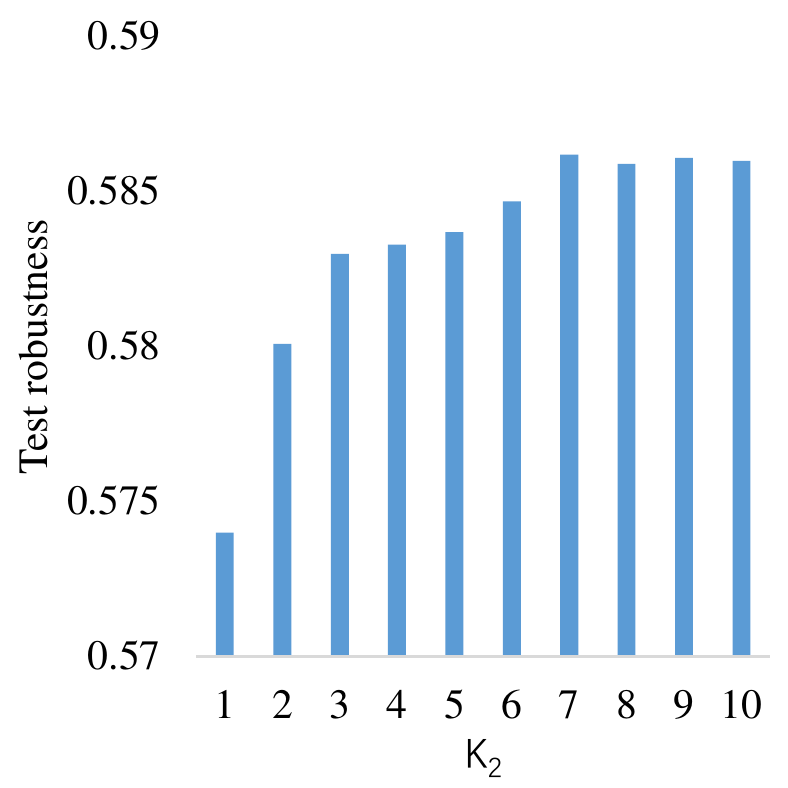}
    }
    \subfigure[Effect of RWP on adversarial robustness and robust overfitting]{
        \includegraphics[width=0.6\columnwidth]{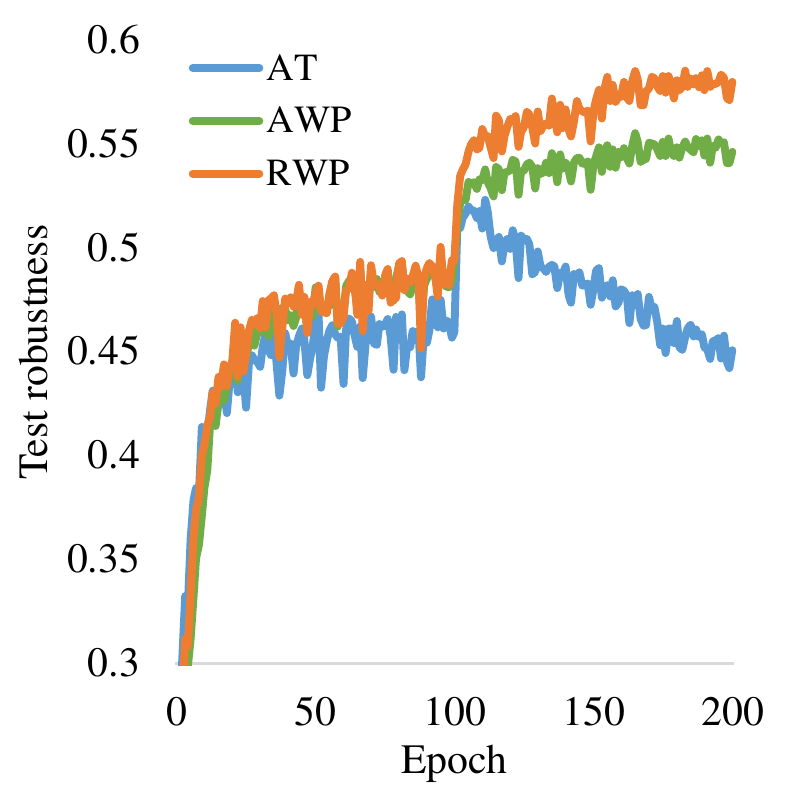}
    }
\caption{The ablation study experiments on CIFAR-10.}
\label{fig:ablation}
\end{figure*}

\subsection{Robustness Evaluation}
\label{robustness evaluation}

\paragraph{Performance Evaluations.} To validate the effectiveness of the proposed RWP, we conduct performance evaluation on vanilla AT, AT-AWP and AT-RWP across different benchmark datasets and threat models using PreAct ResNet-18. We report the accuracy on the test images under PGD-20 attack. The evaluation results are summarized in Table \ref{table:1}, where ``Best'' denotes the highest robustness that ever achieved at different checkpoints and ``Last'' denotes the robustness at the last epoch checkpoint. It is observed vanilla AT suffers from severe robust overfitting (the performance gap between ``best'' and ``last'' is very large). AT-AWP and AT-RWP method  narrow the performance gap significantly over the vanilla AT model due to suppression of robust overfitting. Moreover, on CIFAR-10 dataset under the $L_\infty$ attack, vanilla AT achieves 52.32\% ``best'' test robustness. The AT-AWP approach boosts the performance to 55.54\%. The proposed approach further outperforms both methods by a large margin, improving over vanilla AT by 6.23\%, and is 3.01\% better than AT-AWP, achieving 58.55\% accuracy under the standard 20 steps PGD attack. Similar patten has been observed on other datasets and threat model. AT-RWP consistently improves the test robustness across a wide range of datasets and threat models, demonstrating the effectiveness of the proposed approach.

\paragraph{Benchmarking the state-of-the-art Robustness.} To manifest the full power of our proposed perturbation strategy and also benchmark the state-of-the-art robustness on CIFAR-10 under L$_\infty$ threat model, we conduct experiments on the large capacity network with different baseline methods. We train Wide ResNet-34-10 for AT and TRADES, and Wide ResNet-28-10 for RST following their original papers. We evaluate the adversarial robustness of trained model with various test attack and report the ``best'' test robustness, with the results shown in Table \ref{table:2}. ``Natural'' denotes the accuracy on natural test data. First, it is observed that the natural accuracy of RWP model consistently outperforms AWP by a large margin. It is due to the benefits that our RWP avoids the excessive weight perturbation. Moreover, RWP achieves the best adversarial robustness against almost all types of attack across a wide range of baseline methods, which verifies that RWP is effective in general and improves adversarial robustness reliably rather than improper tuning of hyper-parameters of attacks, gradient obfuscation or masking.

\subsection{Ablation Studies}
\label{Ablation Studies}
In this part, we investigate the impacts of algorithmic components using AT-RWP on PreAct ResNet-18 under $L_\infty$ threat model following the same setting in section \ref{experiment setup}.

\paragraph{The Importance of Minimum Loss Value.} We verify the effectiveness of minimum loss value $c_{min}$, by comparing the performance of models trained using different weight perturbation schemes: 1) AT: standard adversarial training without weight perturbation (equivalent to $c_{min}=0$); 2) AWP: weight perturbation generated via outer maximization in Eq.(\ref{eq1}) (equivalent to $c_{min}=\infty$); 3) RWP: weight perturbation generated using the proposed robust strategy with different $c_{min}$ values. All other hyper-parameters are kept exactly the same other than the perturbation scheme used. The results are summarized in Figure \ref{fig:ablation}(a). It is observed that the test robustness of RWP model first increases and then decreases as the minimum loss value increases, and the best test robustness is obtained at $c_{min}=1.7$. It is evident that RWP with a wide range of $c_{min}$ outperforms both AT and AWP methods, demonstrating its effectiveness. Furthermore, as it is the major component that is different from the AWP pipeline, this result suggests that the proposed LSC constraints is the main contributor to the improved adversarial robustness.

\paragraph{The Impact of Step Number.} We further investigate the effect of step number $K_2$, by comparing the performances of model trained using different perturbation steps. The step number $K_2$ for RWP varies from 1 to 10. The results are shown in Figure \ref{fig:ablation}(b). As expected, when $K_2$ is small, increasing $K_2$ leads higher test robustness. When $K_2$ increases from 7 to 10, the performance is flat, which suggests that the generated weight perturbation is sufficient to comprehensively avoid robust overfitting. Note that extra iterations will not bring computational overhead when the classification loss of adversarial examples exceeds minimum loss value $c_{min}$, as shown in Algorithm \ref{alg:1}. Therefore, we uniformly use $K_2=10$ in our implementation.

\paragraph{Effect on Adversarial Robustness and Robust Overfitting.} We then visualize the learning curves of AT, AWP and RWP, which are summarized in Figure \ref{fig:ablation}(c). It is observed that the test robustness of RWP model continues to increase as the training progresses. In addition, RWP outperforms AWP with a clear margin in the later stage of training. Such observations exactly reflect the nature of our approach which aims to prevent robust overfitting as well as boost the robustness of adversarial training.

%% file: ijcai22_6_conclusion.tex
\section{Conclusion}
In this paper, we proposed a criterion, Loss Stationary Condition (LSC) for constrained weight perturbation. The proposed criterion provides a new understanding of robust overfitting. Based on LSC, we found that elimination of robust overfitting and higher robustness of adversarial training can be achieved by weight perturbation on adversarial examples with small classification loss, rather than adversarial examples with large classification loss. Following this, we proposed a Robust Weight Perturbation (RWP) strategy to regulate the extent of weight perturbation. Comprehensive experiments show that RWP is generic and can improve the state-of-the-art adversarial robustness across different adversarial training approaches, network architectures, threat models and benchmark datasets.